\documentclass{article}

\usepackage{arxiv}
\usepackage[numbers,sort&compress]{natbib}

\usepackage[utf8]{inputenc} 
\usepackage[T1]{fontenc}    
\usepackage{hyperref}       
\usepackage{url}   
\usepackage{algorithm}
\usepackage{booktabs}       
\usepackage{amsfonts}       
\usepackage{nicefrac}       
\usepackage{microtype}      
\usepackage{lipsum}
\usepackage{graphicx}
\usepackage{amsmath} 
\usepackage{amssymb} 
\usepackage{caption}
\usepackage{pgfplots}
\usepackage{tikz}
\usepackage{enumitem}
\usetikzlibrary{arrows.meta, positioning}
\usetikzlibrary{shapes.geometric, arrows.meta, positioning}

\usetikzlibrary{positioning,calc,decorations.pathreplacing}

\usepackage{pgfplotstable}
\pgfplotsset{compat=1.18}

\usepackage[noend]{algpseudocode} 

\usepackage[utf8]{inputenc}

\usepackage{algpseudocode} 

\graphicspath{ {./images/} }

\title{A Long Short-Term Memory (LSTM) Model for Business Sentiment Analysis Based on Recurrent Neural Network \\[0.5em]
\large Published in Sustainable Communication Networks and Application: Proceedings of ICSCN 2020}

\author{
\textbf{Md.\ Jahidul Islam Razin} \\
Department of Computer Science and Engineering \\
University of Asia Pacific \\
\texttt{razin.cse@gmail.com}
\and
\textbf{Md.\ Abdul Karim} \\
Department of Computer Science and Engineering \\
University of Asia Pacific \\
\texttt{karim.cse007@gmail.com}
\and
\textbf{M.\ F.\ Mridha} \\
Department of Computer Science and Engineering \\
Bangladesh University of Business and Technology \\
\texttt{firoz@bubt.edu.bd}
\and
\textbf{S M Rafiuddin} \\
Department of Computer Science and Engineering \\
University of Asia Pacific \\
\texttt{rifat.cse@uap-bd.edu}
\and
\textbf{Tahira Alam} \\
Department of Computer Science and Engineering \\
University of Asia Pacific \\
\texttt{tahira.cse@uap-bd.edu}
}

\begin{document}
\maketitle

\begin{abstract}
Business sentiment analysis (BSA) is one of the significant and popular topics of natural language processing. It is one kind of sentiment analysis techniques for business purposes. Different categories of sentiment analysis techniques like lexicon-based techniques and different types of machine learning algorithms are applied for sentiment analysis on different languages like English, Hindi, Spanish, etc. In this paper, long short-term memory (LSTM) is applied for business sentiment analysis, where a recurrent neural network is used. An LSTM model is used in a modified approach to prevent the vanishing gradient problem rather than applying the conventional recurrent neural network (RNN). To apply the modified RNN model, product review dataset is used. In this experiment, 70\% of the data is trained for the LSTM and the rest 30\% of the data is used for testing. The result of this modified RNN model is compared with other conventional RNN models, and a comparison is made among the results. It is noted that the proposed model performs better than the other conventional RNN models. Here, the proposed model, i.e., the modified RNN model approach has achieved around 91.33\% of accuracy. By applying this model, any business company or e-commerce business site can identify the feedback from their customers about different types of products that customers like or dislike. Based on the customer reviews, a business company or e-commerce platform can evaluate its marketing strategy.
\end{abstract}


\section{Introduction}
Nowadays, the Internet is a fundamental part of our daily life. All fields of information are rising exponentially every moment, where it is easier to share our opinions on e-commerce websites, forums and media for various types of products and services. It provides essential information about different domains and social applications. It is difficult to handle these huge amounts of data manually. For this reason, business sentiment analysis (BSA) is very feasible, which also provides an idea of the requirements of people. It has become a popular research topic in the natural language processing domain. Analysis of reviews can quickly extract information from a text and can also define the target and opinion polarity. Various types of social applications and websites can use BSA to forecast consumer patterns, economic policies and stock market movements.

Researchers perform business sentiment analysis using various machine learning (ML) techniques. ML’s support learning models include support vector machines (SVM) \cite{Vishwanathan2002}, logistic regression \cite{Wright1995}, naive Bayes \cite{Zhang2004}, random forests \cite{Breiman2001} and so on. Artificial neural network (ANN) is one of the areas of ML \cite{AgatonovicKustrin2000}. ANN also has various forms like recurrent neural network (RNN) \cite{Mikolov2010} and convolutional neural network (CNN) \cite{Huang2014}. Artificial neural networks are mainly constructed using three layers: the input layer, hidden layer and output layer. This concept is extended in deep learning. Deep learning is constructed using more than two hidden layers; the depth of the network is defined by the number of layers used in the hidden part. Deep learning is a part of artificial neural networks and gives far better performance than other ML techniques such as SVM or NB. These techniques are used in all natural language processing fields like speech, entity and pattern recognition \cite{Liang2013} and in computer vision techniques. For its accuracy, deep learning has become very popular. Good accuracy can be achieved if the representation of data is perfect. Conventional ML algorithms depend on handcrafted features, whereas deep learning requires high computational ability and storage to increase the number of hidden layers compared to conventional ML algorithms, thereby achieving better performance. Deep learning techniques can adapt more swiftly as they receive more training data. As business sentiment analysis needs to process large volumes of data for prediction, this motivated us to apply deep learning techniques for business sentiment analysis \cite{Bhatia2018}.

The importance of consumer views, the vast amounts of public-centric data, and the unpredictability of the market climate have led organizations to introduce monitoring and tracking measures such as sentiment analysis, which is here performed as business sentiment analysis. 

The principal contribution of the proposed research work can be summed up as follows. This paper presents the study of the business sentiment function by a list of product feedback. Using the proposed model, the reviews are divided into three categories: positive, neutral and negative. A well-defined text dataset is used to apply long short-term memory (LSTM) as a part of a modified RNN model that gives better accuracy results in comparison to conventional RNN models. This is a different approach when compared to traditional feed-forward networks \cite{Schmidt1992}. 

A feed-forward network takes a limited text for predicting the next word. Here, RNN can easily use previous words for prediction. An RNN model sees the text as a signal made up of terms, where it is presumed that recursive weights reflect short-term memory. LSTM is different, as it is part of the RNN structure in the ANN network: it accommodates both long- and short-term patterns and resolves the vanishing gradient problem. Thus, LSTM is now approved in a number of applications and is a promising practice for business sentiment analysis.

This article is structured as follows. Section 2 explains the related works and motivation of this research. Section 3 discusses the methodology. Section 4 describes the implemented tools. Section 5 presents results and analysis. Section 6 concludes the paper.

\section{Related Work}
Several studies have been performed on sentiment analysis for different languages using various neural network architectures. Xu et al.\ \cite{Xu2014} applied supervised learning algorithms—perceptron, Naive Bayes, and support vector machine—to predict reviewer ratings, using 70\% of the data for training and 30\% for testing, and reported precision and recall values for each classifier. Deep learning has also become popular in sentiment analysis. Rani and Kumar \cite{Rani2019} applied a convolutional neural network (CNN) for sentiment analysis, using 50\% of the data for training and 50\% for testing after cross-validation. Mikolov et al.\ \cite{Mikolov2010} proposed a recurrent neural network (RNN) model for processing sequential text data, wherein the hidden state at each time step incorporates information from the previous step, effectively forming a loop over time to capture sequential dependencies.

However, the standard RNN model can suffer from vanishing gradients and thus struggles to capture long-term dependencies when there are large intervals between related words. The back-propagation through time (BPTT) optimization exacerbates this issue by propagating error gradients through many time steps, leading to information loss. To address these limitations, we employ long short-term memory (LSTM) networks—which introduce gated mechanisms to preserve both long- and short-term patterns—thereby mitigating the vanishing gradient problem and improving performance in business sentiment analysis.

\section{Methodology}
This section describes our method for business sentiment analysis using reviewers’ text data. We employ long short-term memory (LSTM) networks, which are designed to avoid the vanishing gradient problem and capture long-term dependencies by replacing each RNN node with an LSTM cell that can store and recall information over long sequences. As a result, LSTM-based sentiment analysis outperforms traditional RNN approaches for business applications.

Our proposed method is divided into four phases:
\begin{enumerate}
  \item Discussion about RNN and LSTM architecture.
  \item Data processing.
  \item Training the model.
  \item Testing new data.
\end{enumerate}

\subsection{Discussion About RNN and LSTM Architecture}
LSTM is a special case of RNN and there are many similarities between them. This work presents structural diagrams and a brief discussion of both RNN and LSTM architectures.

\begin{figure}[ht]
  \centering
  \includegraphics[width=0.8\textwidth]{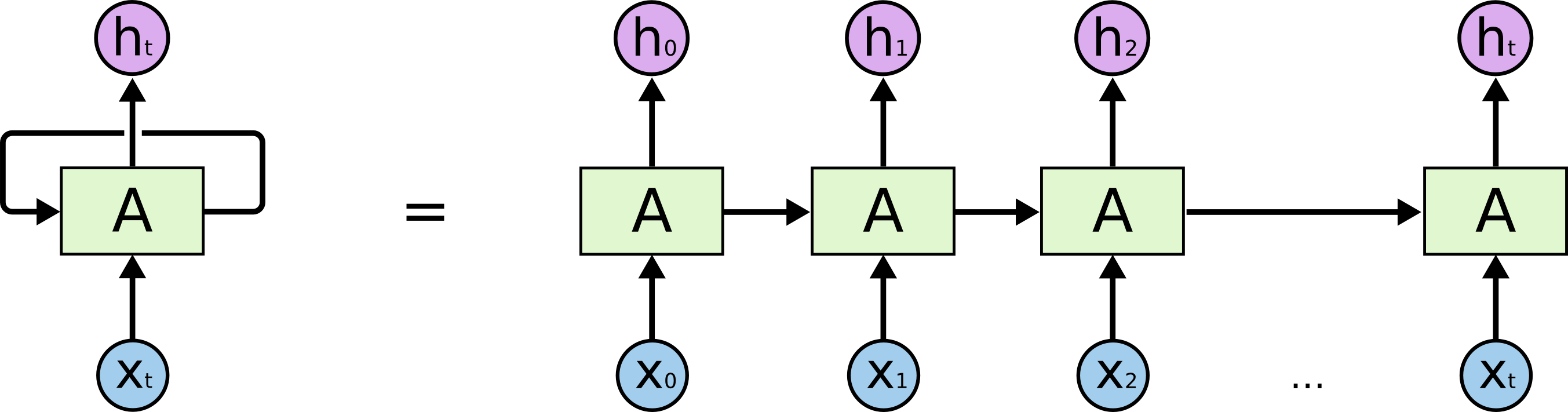}
  \caption{The structural diagram of the RNN model.}
  \label{fig:rnn_architecture}
\end{figure}

Figure~\ref{fig:rnn_architecture} \cite{Graves2012} illustrates a traditional RNN model where $X(t)$ is the input, $h(t)$ is the output, and $A$ is the recurrent weight matrix that propagates information from one time step to the next. One output feeds back into the network, passing information forward through time.

\begin{figure}[ht]
  \centering
  \includegraphics[width=0.8\textwidth]{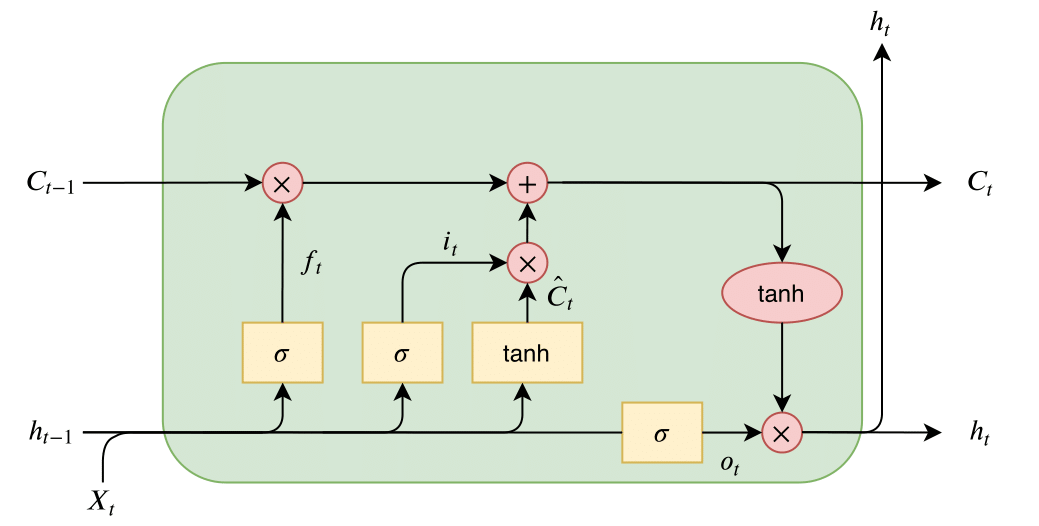}
  \caption{Structural diagram of an LSTM cell.}
  \label{fig:lstm_cell1}
\end{figure}

\begin{figure}[ht]
  \centering
  \includegraphics[width=0.8\textwidth]{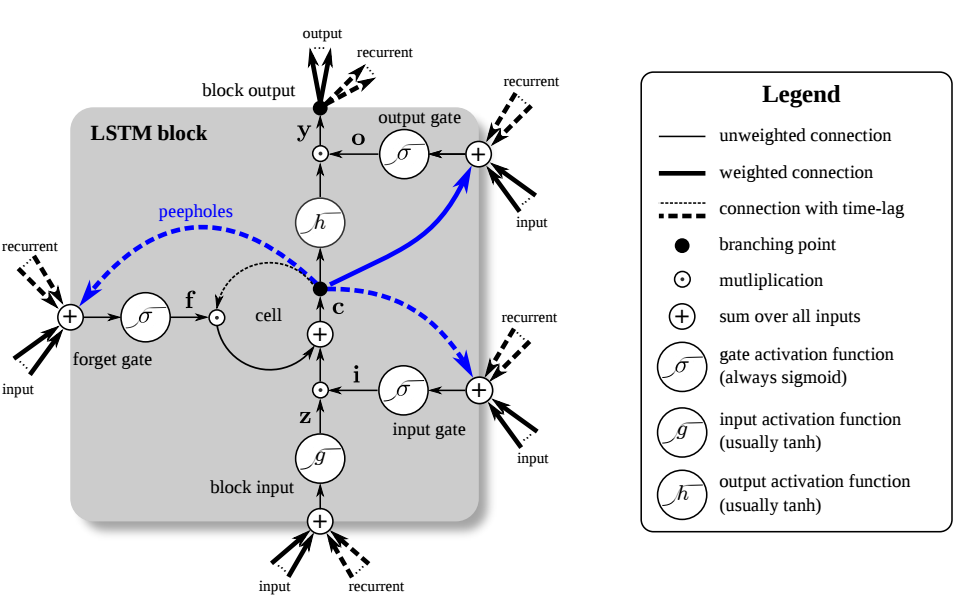}
  \caption{Structural diagram of the LSTM cell.}
  \label{fig:lstm_cell2}
\end{figure}

Figure~\ref{fig:lstm_cell1} \cite{Graves2012} shows a single LSTM cell. The LSTM architecture is similar to the standard RNN, except that the hidden layer is replaced by a memory block composed of specialized gates \cite{Graves2012}. In these diagrams, the symbols have the following meanings: $X$ scales input information; $\sigma$ denotes a sigmoid layer; $\tanh$ denotes a hyperbolic tangent layer; $h(t-1)$ is the previous hidden state; $c(t-1)$ is the previous cell state; $X(t)$ is the current input; $c(t)$ is the updated cell state; and $h(t)$ is the current hidden state.

Standard RNNs use only a $\tanh$ activation to help mitigate the vanishing gradient problem \cite{Zhang2015}, but they cannot selectively forget or remember information. LSTM adds sigmoid-based “forget” and “input” gates alongside $\tanh$ layers to control what information is discarded or stored.

The LSTM computation proceeds in four main steps \cite{GravesChapter2012,Soutner2013,Sundermeyer2014a,Sundermeyer2015}:
\begin{enumerate}
  \item Compute forget gate and input gate activations.
  \item Update the cell state using the input and forget gates.
  \item Compute the output gate activation.
  \item Produce the new hidden state and finalize the updated cell state.
\end{enumerate}

All equations are shown below \cite{Graves2012}.

\textbf{Input Gate:}
\begin{equation}
i_t = \sigma\bigl(x_t U^i + h_{t-1} W^i\bigr)
\end{equation}

\textbf{Forget Gate:}
\begin{equation}
f_t = \sigma\bigl(x_t U^f + h_{t-1} W^f\bigr)
\end{equation}

\textbf{Cell Candidate:}
\begin{equation}
\tilde{C}_t = \tanh\bigl(x_t U^g + h_{t-1} W^g\bigr)
\end{equation}

\textbf{Cell State Update:}
\begin{equation}
C_t = f_t \times C_{t-1} + i_t \times \tilde{C}_t
\end{equation}

\textbf{Output Gate:}
\begin{equation}
o_t = \sigma\bigl(x_t U^o + h_{t-1} W^o\bigr)
\end{equation}

\textbf{Hidden State:}
\begin{equation}
h_t = \tanh(C_t) \times o_t
\end{equation}

\begin{figure}[ht]
  \centering
  \includegraphics[width=0.8\textwidth]{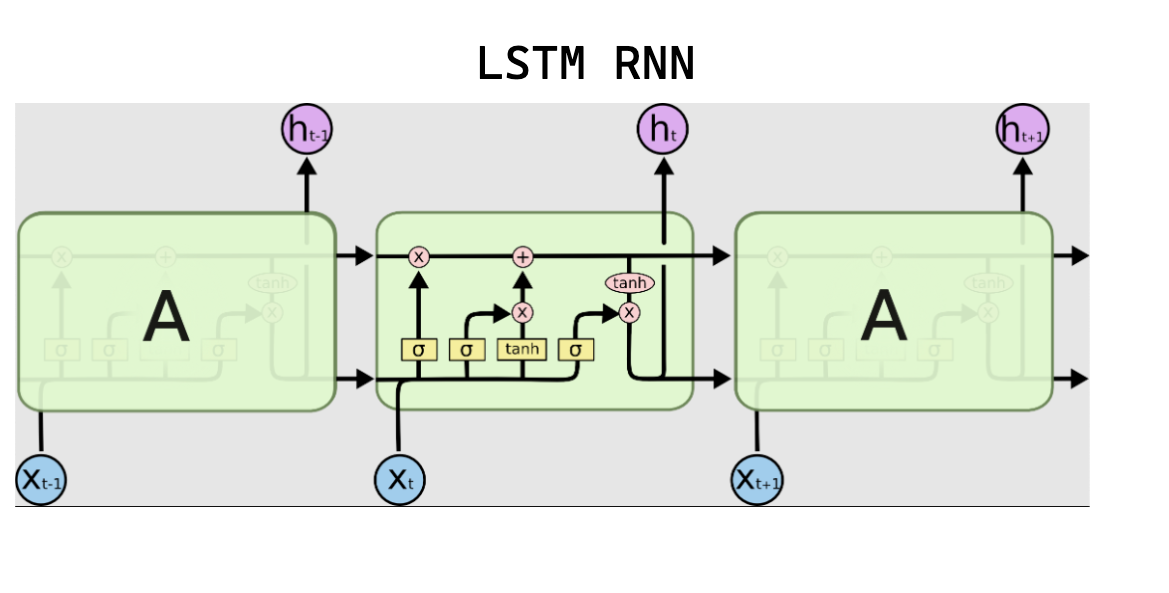}
  \caption{LSTM architecture.}
  \label{fig:lstm_architectu}
\end{figure}

\begin{table}[ht]
  \centering
  \caption{Model Summary}
  \label{tab:model_summary}
  \begin{tabular}{l l r}
    \hline
    \textbf{Layer (type)} & \textbf{Output Shape} & \textbf{Param \#} \\
    \hline
    embedding\_1 (Embedding) & (None, None, 32) & 640\,000 \\
    lstm\_1 (LSTM)           & (None, 100)      & 53\,200  \\
    dense\_1 (Dense)         & (None, 1)        & 101      \\
    \hline
    \multicolumn{2}{l}{Total params:}             & 693\,301 \\
    \multicolumn{2}{l}{Trainable params:}         & 693\,301 \\
    \multicolumn{2}{l}{Non-trainable params:}     & 0        \\
    \hline
  \end{tabular}
\end{table}

\begin{figure}[ht]
  \centering
  \includegraphics[width=0.8\textwidth]{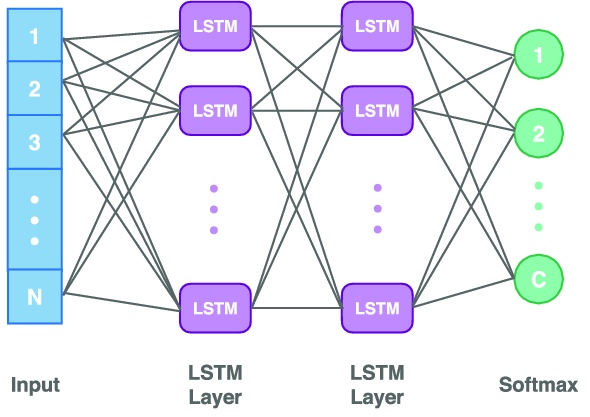}
  \caption{LSTM architecture}
  \label{fig:lstm_architecture_6}
\end{figure}

RNN and LSTM consider the problem of the vanishing gradient when processing a sequence of text. While standard RNNs can exhibit lower error rates on short sequences, they struggle with long-range dependencies. LSTM, by contrast, is more effective at overcoming the vanishing gradient problem through its gated memory cells. In our business sentiment analysis, we apply the modified RNN-based LSTM model. The proposed model architecture is shown in Table~\ref{tab:model_summary} and Figure~\ref{fig:lstm_architecture_6} \cite{Graves2012}.

\subsection{Data Processing}
To train the proposed model, a business review dataset is collected from amazon.com on product analysis. The dataset was created by Web scraping or APIs. Researchers compiled the information from Amazon Review Information (ARD) when it comes to Amazon datasets \cite{Abadi2016}. This dataset contained punctuation and HTML tags. In order to vanish these, a function is used for taking a text string as a parameter and then preprocess the string to remove punctuation and HTML tags from the given string. These punctuation and HTML tags are replaced with an empty space. All the single characters and multiple spaces are removed. At last for each term, the analysis text is naturally divided into emotions of the business class. Then, a tool is used for vector representation to transform every term into a vector using the vector equation:
\begin{equation}
v \in \mathbb{R}^{1 \times d}
\end{equation}
which is known as the word embedding. Here, word2vector tools are used. Then the cycle was followed by the natural order; the sentence of participles was traversed from left to right, which is the forward calculation of LSTM. The results of the production depended on the probabilities of the word at ‘t’ time and also the sequence of vocabulary giving before ‘t’ time. Finally, the error was determined by the likelihood of a common distribution of all the words in the sentences.

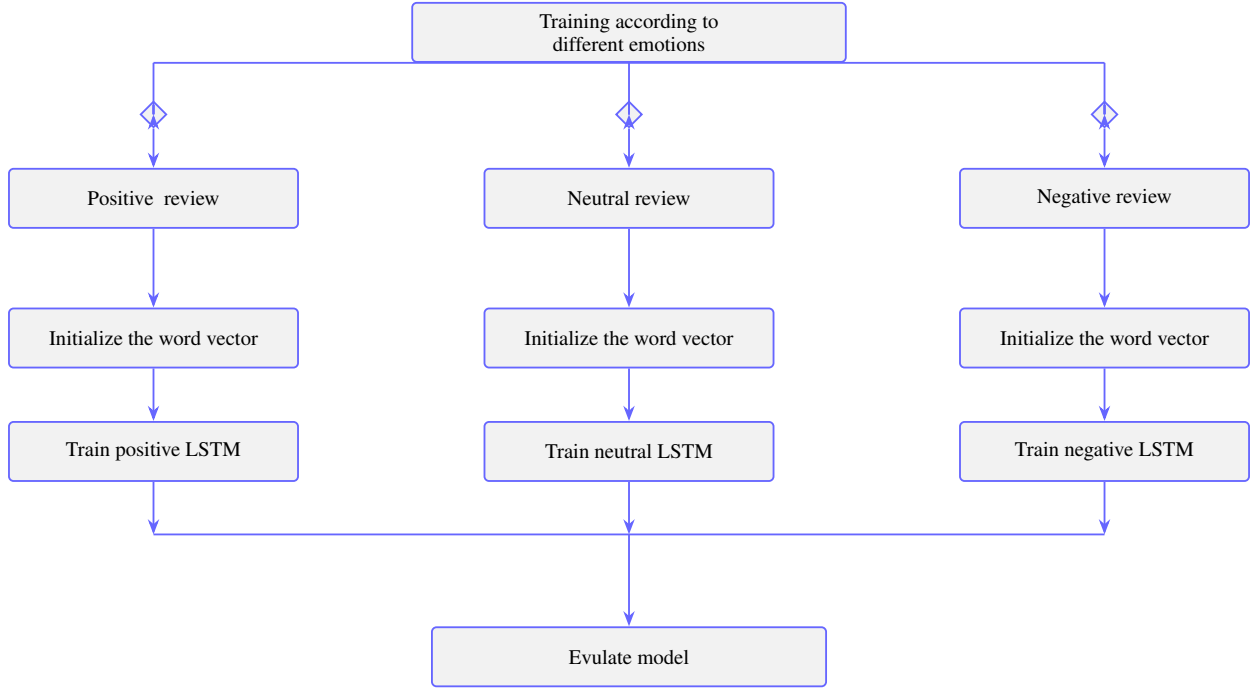
\begin{figure}[ht]
\centering
\resizebox{\linewidth}{!}{%
\begin{tikzpicture}[
  >=Stealth,
  node distance=8mm and 20mm,
  every node/.style={font=\small},
  box/.style={draw=blue!60, rounded corners=2pt, thick, fill=gray!10,
              minimum width=43mm, minimum height=9mm, align=center},
  midbox/.style={box, minimum width=44mm},
  smallbox/.style={box, minimum width=44mm},
  trainbox/.style={box, minimum width=44mm},
  evalbox/.style={box, minimum width=60mm},
  split/.style={diamond, draw=blue!60, thick, fill=gray!10, inner sep=0pt,
                minimum size=4mm},
  arr/.style={->, draw=blue!60, thick}
]

\node (title) [box, minimum width=66mm] {Training according to\\different emotions};

\node (L0) [below left=16mm and 38mm of title] {};
\node (C0) [below=16mm of title] {};
\node (R0) [below right=16mm and 38mm of title] {};

\node (Ls) [split, above=6mm of L0] {};
\node (Cs) [split, above=6mm of C0] {};
\node (Rs) [split, above=6mm of R0] {};

\draw[draw=blue!60, thick] (title.south -| Ls) -- (title.south -| Rs);
\draw[arr] (title.south -| Ls) |- (Ls);
\draw[arr] (title.south -| Cs) |- (Cs);
\draw[arr] (title.south -| Rs) |- (Rs);

\node (pos) [smallbox, below=6mm of Ls] {Positive\ \ review};
\node (neu) [smallbox, below=6mm of Cs] {Neutral review};
\node (neg) [smallbox, below=6mm of Rs] {Negative review};

\draw[arr] (Ls) -- (pos);
\draw[arr] (Cs) -- (neu);
\draw[arr] (Rs) -- (neg);

\node (pinit) [midbox, below=12mm of pos] {Initialize the word vector};
\node (ninit) [midbox, below=12mm of neu] {Initialize the word vector};
\node (ginit) [midbox, below=12mm of neg] {Initialize the word vector};

\draw[arr] (pos) -- (pinit);
\draw[arr] (neu) -- (ninit);
\draw[arr] (neg) -- (ginit);

\node (ptrain) [trainbox, below=8mm of pinit] {Train positive LSTM};
\node (ntrain) [trainbox, below=8mm of ninit] {Train neutral LSTM};
\node (gtrain) [trainbox, below=8mm of ginit] {Train negative LSTM};

\draw[arr] (pinit) -- (ptrain);
\draw[arr] (ninit) -- (ntrain);
\draw[arr] (ginit) -- (gtrain);

\coordinate (mergeL) at ($(ptrain.south)+(0,-8mm)$);
\coordinate (mergeC) at ($(ntrain.south)+(0,-8mm)$);
\coordinate (mergeR) at ($(gtrain.south)+(0,-8mm)$);

\draw[arr] (ptrain.south) -- (mergeL);
\draw[arr] (ntrain.south) -- (mergeC);
\draw[arr] (gtrain.south) -- (mergeR);

\draw[draw=blue!60, thick]
  (mergeL) -- ($(mergeL)!0.5!(mergeR)$)
  -- (mergeR);

\node (eval) [evalbox, below=14mm of mergeC] {Evulate model};

\draw[arr] ($(mergeL)!0.5!(mergeR)$) -- (eval);

\end{tikzpicture}%
}
\caption{Business sentimental emotion classification}
\label{fig:sentiment}
\end{figure}

\begin{figure}[ht]
\centering
\resizebox{\linewidth}{!}{%
\begin{tikzpicture}[
  >=Stealth,
  node distance=6mm and 12mm,
  every node/.style={font=\small},
  num/.style   ={draw=blue!60, thick, fill=blue!10, rounded corners=2pt,
                  minimum width=16mm, minimum height=7mm, align=center},
  blk/.style   ={draw=blue!60, thick, fill=blue!10, rounded corners=2pt,
                  minimum width=22mm, minimum height=7.5mm, align=center},
  lstmblk/.style={blk, minimum width=20mm},
  fc/.style    ={draw=blue!60, thick, fill=blue!15, rounded corners=2pt,
                  minimum width=98mm, minimum height=10mm, align=center},
  outbox/.style={blk, minimum width=18mm},
  dot/.style   ={circle, fill=blue!60, inner sep=1.8pt},
  arr/.style   ={->, very thick, draw=blue!60}
]

\node (t1t) at (0,0) {This};
\node (t2t) [right=22mm of t1t] {Product};
\node (t3t) [right=22mm of t2t] {is};
\node (t4t) [right=22mm of t3t] {Awesome};

\node (n1) [num, below=2mm of t1t] {24};
\node (n2) [num, below=2mm of t2t] {355};
\node (n3) [num, below=2mm of t3t] {18};
\node (n4) [num, below=2mm of t4t] {360};

\node (toklab) [right=9mm of t4t, anchor=west] {Tokenize};

\node (e1) [blk, below=8mm of n1] {Embedding};
\node (e2) [blk, below=8mm of n2] {Embedding};
\node (e3) [blk, below=8mm of n3] {Embedding};
\node (e4) [blk, below=8mm of n4] {Embedding};

\foreach \a/\b in {n1/e1,n2/e2,n3/e3,n4/e4}{
  \draw[arr] (\a) -- (\b);
}

\node (l1) [lstmblk, below=7mm of e1] {LSTM};
\node (l2) [lstmblk, below=7mm of e2] {LSTM};
\node (l3) [lstmblk, below=7mm of e3] {LSTM};
\node (l4) [lstmblk, below=7mm of e4] {LSTM};

\foreach \a/\b in {e1/l1,e2/l2,e3/l3,e4/l4}{
  \draw[arr] (\a) -- (\b);
}

\draw[arr] (l1.east) -- (l2.west);
\draw[arr] (l2.east) -- (l3.west);
\draw[arr] (l3.east) -- (l4.west);

\node (d1) [dot, below=4.5mm of l1] {};
\node (d2) [dot, below=4.5mm of l2] {};
\node (d3) [dot, below=4.5mm of l3] {};
\node (d4) [dot, below=4.5mm of l4] {};

\foreach \a in {1,2,3,4}{
  \draw[arr] (l\a.south) -- (d\a);
}

\node (fc) [fc, below=6mm of d2, anchor=center, xshift=11mm] {Fully Connected Layer};

\node (fd1) [dot, below=6mm of fc.west, xshift=16mm] {};
\node (fd2) [dot, below=6mm of fc.west, xshift=16mm+22mm] {};
\node (fd3) [dot, below=6mm of fc.west, xshift=16mm+44mm] {};
\node (fd4) [dot, below=6mm of fc.east, xshift=-16mm] {};

\draw[arr] (d1) -- (fc.north -| d1);
\draw[arr] (d2) -- (fc.north -| d2);
\draw[arr] (d3) -- (fc.north -| d3);
\draw[arr] (d4) -- (fc.north -| d4);

\foreach \pt in {fd1,fd2,fd3,fd4}{
  \draw[arr] (fc.south -| \pt) -- (\pt);
}

\node (s1) [blk, below=7mm of fd1] {Sigmoid};
\node (s2) [blk, below=7mm of fd2] {Sigmoid};
\node (s3) [blk, below=7mm of fd3] {Sigmoid};
\node (s4) [blk, below=7mm of fd4] {Sigmoid};

\foreach \a/\b in {fd1/s1,fd2/s2,fd3/s3,fd4/s4}{
  \draw[arr] (\a) -- (\b);
}

\node (out) [outbox, below=10mm of s4] {Output};
\draw[arr] (s4) -- (out);

\foreach \a/\b in {t1t/n1,t2t/n2,t3t/n3,t4t/n4}{
  \draw[blue!60, thick] (\a.south) -- (\b.north);
}

\node (ps) [below=7mm of s2] {Probability score};

\end{tikzpicture}%
}
\caption{Tokenized LSTM pipeline with embeddings, recurrent layer, fully connected layer, sigmoid activations, and output probability.}
\label{fig:lstm-pipeline}
\end{figure}
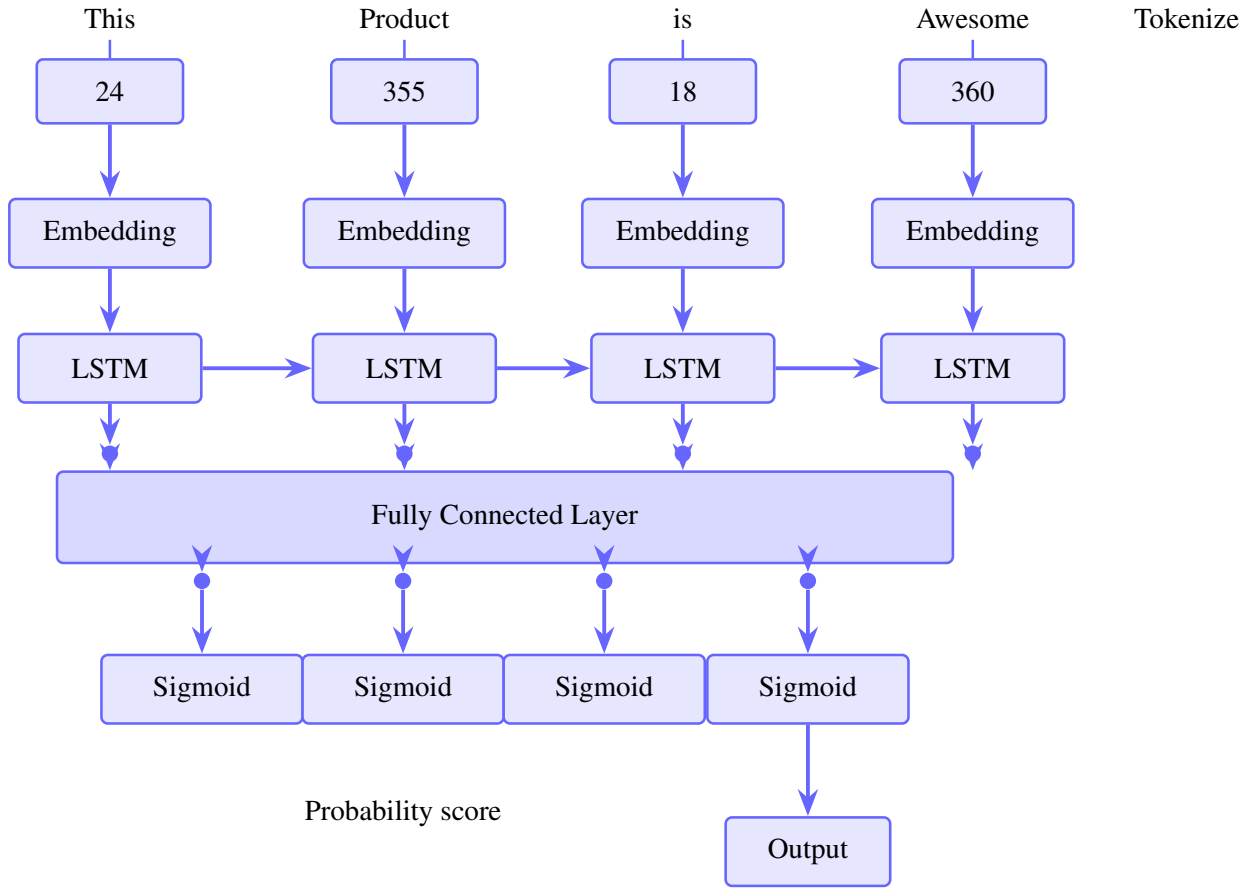

\subsection{Training the Model}
\label{subsec:training_model}

After preprocessing the data and splitting the dataset, LSTM model is used to train
our model. LSTM layer is created with 100 neurons and also added a dense layer
with sigmoid activation function.

The process is introduced as follows:
\begin{itemize}
  \item Due to their emotional marks, all training data is divided into three groups, has positive, negative and neutral. The LSTM models are then trained in each data category and with multiple LSTM models resulting in them as well. This is done for equal ratings.
  \item To get a new input review, the LSTM models are available in the training phase evaluated on the new input review. The model which is giving the smallest error value is assigned to the new input review.
\end{itemize}

Figure~\ref{fig:sentiment} is showing the structure of the processing of the training phase: This
model could overcome the vanishing gradient problem completely than the conventional RNN model. It also better performs in many experiments, like as structure
with conjunctions, such as, ‘not only…but also…,’ ‘However,’‘in addition,’ etc.

\subsection{Testing the Model}
After training the model, new text data is used for testing. When a new
product review arrives, the model classifies it as negative, positive, or neutral. 
As shown in Figure~\ref{fig:lstm-pipeline}, this illustrates the working process 
of our RNN-LSTM model in business sentiment analysis.

\section{Implementation Tools}
The proposed work has used the Jupyter Notebook \cite{Perkel2018}, which is an open-source web platform. It has helped to develop an environment to perform our experiments and also create and share all types of documents like code, text, equations and visualizations. The LSTM-RNN model is developed using TFLearn (deep learning library) Python packages, which are installed on top of TensorFlow \cite{Abadi2016}. The proposed work also uses Keras (NNs API) \cite{Gulli2017}, which is written in Python and runs on TensorFlow.

\section{Result and Analysis}
The analysis of the outcome along with the result for the entire system is discussed in this section. The outcome is split into three parts.

\begin{table}[ht]
  \centering
  \caption{Accuracy of LSTM model}
  \label{tab:accuracy_lstm}
  \begin{tabular}{c c c}
    \hline
    \textbf{Epoch} & \textbf{Training accuracy} & \textbf{Testing accuracy} \\
    \hline
    10 & 0.9504 & 0.8954 \\
    30 & 0.9551 & 0.9092 \\
    50 & 0.9623 & 0.9193 \\
    \hline
  \end{tabular}
\end{table}

\subsection{Experimental Dataset}
\label{subsec:experimental_dataset}
Amazon Review Information Dataset (ARD) \cite{He2016} contains 142.8 million ratings and
a variety of metadata. This work has taken a selected amount of data from this dataset.
After taking the data from dataset, all the data has been categorized into positive,
neutral and negative. Then Word2Vec tools are used to create word embedding for
our LSTM model \cite{Mikolov2013}. After that, it has been gone through the LSTM layer \cite{Graves2012}. There is a
\texttt{Dense()} layer, which is the final layer of the model \cite{Gulli2017}. It can crunch all the output that
is coming from the LSTM layer and convert it as a single numeric value of 0.0 and
1.0.

Total of 25,000 product reviews, which are divided into a 70\% items for the training
set and a 30\% items for the testing set. Then train and test have been performed by
the proposed model. It achieved 95.04\% (from Table~\ref{tab:accuracy_lstm}) accuracy on the training data
and 89.54\% accuracy on the test data for 10 epochs. Then increasing the epoch size
into 20, our accuracy for training data 95.51\% and accuracy for test data 90.92\%.
Again, the epoch size is increased from 20 to 50, then the accuracy on training data
will be 96.23\% and accuracy on testing the data will be 91.33\%. If epoch size is
increased more than 50 iteration, the deference between training and test result is
increased a lot and overfitted. That is why the epoch size is not increased more than
50. Figure~\ref{fig:loss_curves} is showing the training and testing loss; after 20 iteration, here blue line
are showing training loss and orange line are showing testing loss.

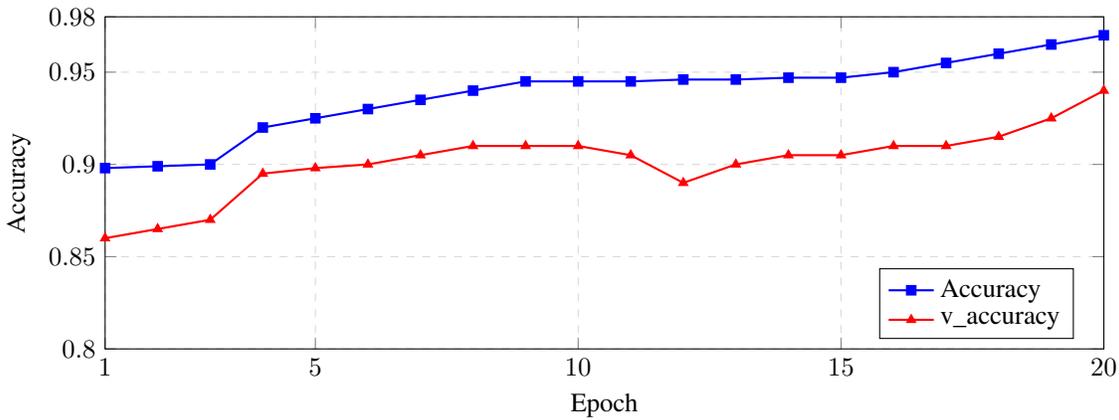
\begin{figure}[ht]
  \centering
  \begin{tikzpicture}
    \begin{axis}[
      width=0.9\linewidth,
      height=6cm,
      xlabel={Epoch},
      ylabel={Accuracy},
      xmin=1, xmax=20,
      ymin=0.80, ymax=0.98,
      xtick={1,5,10,15,20},
      ytick={0.80,0.85,0.90,0.95,0.98},
      grid=both,
      grid style={gray!30, dashed},
      legend pos=south east,
      legend cell align=left
    ]

    \addplot[
      blue, thick,
      mark=square*, mark options={scale=0.8,fill=blue}
    ] coordinates {
      (1,0.898) (2,0.899) (3,0.900) (4,0.920) (5,0.925)
      (6,0.930) (7,0.935) (8,0.940) (9,0.945) (10,0.945)
      (11,0.945) (12,0.946) (13,0.946) (14,0.947) (15,0.947)
      (16,0.950) (17,0.955) (18,0.960) (19,0.965) (20,0.970)
    };
    \addlegendentry{Accuracy}

    \addplot[
      red, thick,
      mark=triangle*, mark options={scale=0.8,fill=red}
    ] coordinates {
      (1,0.860) (2,0.865) (3,0.870) (4,0.895) (5,0.898)
      (6,0.900) (7,0.905) (8,0.910) (9,0.910) (10,0.910)
      (11,0.905) (12,0.890) (13,0.900) (14,0.905) (15,0.905)
      (16,0.910) (17,0.910) (18,0.915) (19,0.925) (20,0.940)
    };
    \addlegendentry{v\_accuracy}

    \end{axis}
  \end{tikzpicture}
  \caption{Training and validation accuracy over 20 epochs. Blue squares indicate training accuracy and red triangles indicate validation accuracy.}
  \label{fig:accuracy_curves}
\end{figure}

Figure~\ref{fig:accuracy_curves} is showing the training and testing accuracy; here, blue line is indicating
training accuracy and orange line is indicating testing accuracy. Our model test data
accuracy is 91.33\%.

\begin{figure}[ht]
  \centering
  \begin{tikzpicture}
    \begin{axis}[
      width=0.9\linewidth,
      height=6cm,
      xlabel={Epoch},
      ylabel={Error},
      xmin=1, xmax=20,
      ymin=0.15, ymax=0.30,
      xtick={1,5,10,15,20},
      ytick={0.15,0.20,0.25,0.30},
      grid=both,
      grid style={gray!30, dashed},
      legend pos=north east,
      legend cell align=left
    ]

    \addplot[
      blue, thick,
      mark=square*, mark options={scale=0.8,fill=blue}
    ] coordinates {
      (1,0.210) (2,0.190) (3,0.180) (4,0.178) (5,0.177)
      (6,0.176) (7,0.175) (8,0.174) (9,0.174) (10,0.173)
      (11,0.172) (12,0.170) (13,0.169) (14,0.168) (15,0.167)
      (16,0.166) (17,0.165) (18,0.164) (19,0.164) (20,0.163)
    };
    \addlegendentry{loss}

    \addplot[
      red, thick,
      mark=triangle*, mark options={scale=0.8,fill=red}
    ] coordinates {
      (1,0.250) (2,0.230) (3,0.210) (4,0.210) (5,0.208)
      (6,0.205) (7,0.202) (8,0.200) (9,0.200) (10,0.220)
      (11,0.210) (12,0.195) (13,0.190) (14,0.185) (15,0.184)
      (16,0.182) (17,0.180) (18,0.178) (19,0.177) (20,0.175)
    };
    \addlegendentry{valid\_loss}

    \end{axis}
  \end{tikzpicture}
  \caption{Training and validation loss over 20 epochs. Blue squares indicate training loss and red triangles indicate validation loss.}
  \label{fig:loss_curves}
\end{figure}

\begin{figure}[H]
\centering
\begin{tikzpicture}
\begin{axis}[
  ybar,
  bar width=16pt,
  enlarge x limits=0.25,
  symbolic x coords={KNN,SVM,SVM + Naive Bayes,Our model(LSTM)},
  xtick=data,
  ymin=0, ymax=100,
  xlabel={Model},
  ylabel={Accuracy (\%)},
  grid=both,
  grid style={gray!35},
  nodes near coords,
  nodes near coords style={font=\small, /pgf/number format/fixed},
  nodes near coords align={vertical},
  tick align=inside,
  width=\linewidth,
  height=0.55\linewidth,
  legend style={at={(0.5,-0.15)},anchor=north,legend columns=-1},
]
  \addplot+[draw=black, fill=blue!70]   coordinates {(KNN,62) (SVM,0) (SVM + Naive Bayes,0) (Our model(LSTM),0)};
  \addplot+[draw=black, fill=green!70]  coordinates {(KNN,0) (SVM,82) (SVM + Naive Bayes,0) (Our model(LSTM),0)};
  \addplot+[draw=black, fill=red!70]    coordinates {(KNN,0) (SVM,0) (SVM + Naive Bayes,56) (Our model(LSTM),0)};
  \addplot+[draw=black, fill=orange!85] coordinates {(KNN,0) (SVM,0) (SVM + Naive Bayes,0) (Our model(LSTM),88)};
  \legend{KNN, SVM, SVM + Naive Bayes, Our model (LSTM)}
\end{axis}
\end{tikzpicture}
\caption{Comparison with other models (Accuracy).}
\label{fig:comparison_models}
\end{figure}
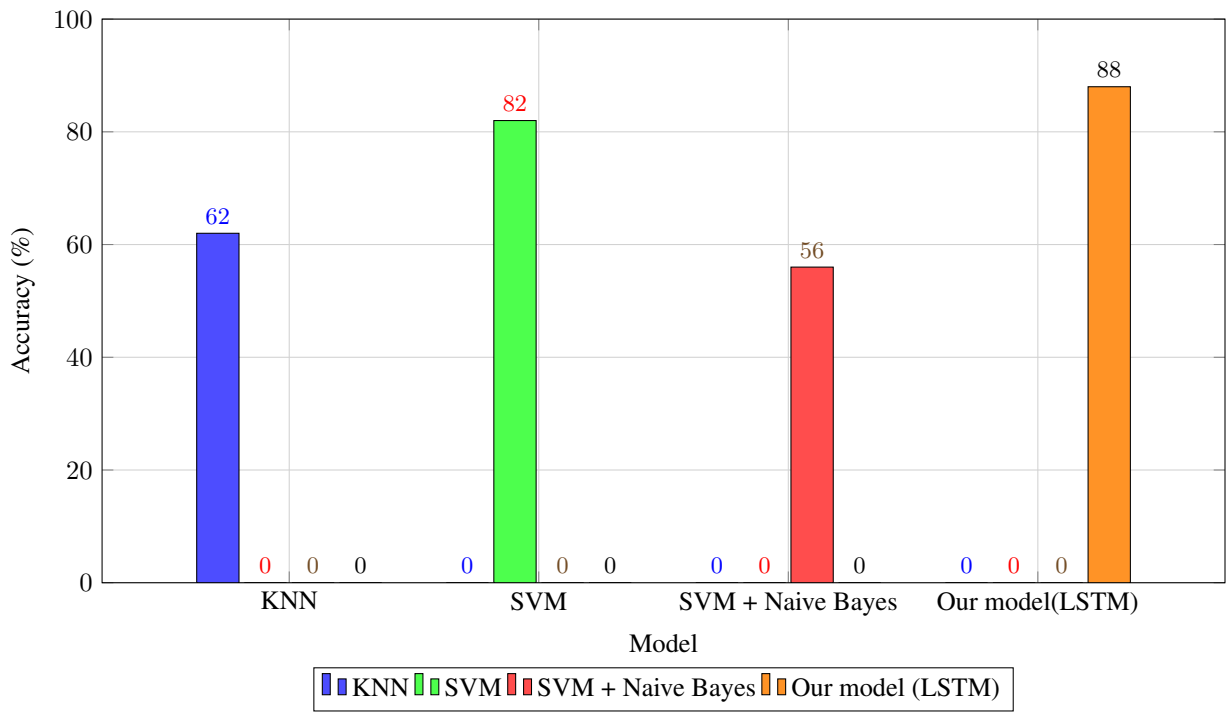

\begin{table}[ht]
  \centering
  \caption{Review type}
  \label{tab:review_type}
  \begin{tabular}{l l}
    \hline
    \textbf{Prediction}           & \textbf{Result} \\
    \hline
    Greater than 0.80             & Excellent       \\
    Greater than 0.60             & Better          \\
    Greater than 0.50             & Good            \\
    Less than 30                  & Very bad        \\
    Greater than 0.30             & Bad             \\
    \hline
  \end{tabular}
\end{table}

\subsection{Experiment Results Analysis}
After training and testing this model, it can easily classify a new product review, which is previously unseen. Previously unseen product review of, “The product was not so great.” Here, the prediction probability value is around 0.1368. From Table~2: When the value is less than 0.30, the model predicts it as very bad. If the value is greater than 0.30, then it is predicted as bad. When the value is greater than 0.50, then it is positive. When the value is greater than 0.60, then it is predicted as good. When it is greater than 0.80, then it is predicted as excellent. For multi-valued analysis, it is required to encode as positive (0, 0, 1), neutral (0, 1, 0) and negative (1, 0, 0).

\section{Comparison}
Mohammad Rezwanul Huq and Ahmad Ali proposed two models for sentiment classification \cite{Huq2017}. Their first model is based on k-nearest neighbor (KNN) and the second on support vector machine (SVM). They achieved 84.32\% accuracy with the KNN model and 67.03\% accuracy with the SVM model. Xu et al.\ \cite{Xu2014} applied SVM and Naive Bayes to predict reviewer ratings, achieving 88.8\% accuracy on training data and 59.1\% accuracy on testing data. Figure~\ref{fig:comparison_models} shows a comparison of different models and their accuracies. Our model achieved 91.33\% accuracy, outperforming all other proposed models.

\section{Conclusion}
In this paper, an LSTM model based on RNN is used to evaluate the business sentiment. It covers all the sequences and performance of business sentiment analysis. This model is used for multi-classification in product reviews. This model can be used for any kind of product review datasets for different languages. In the future, it will be attempted to improve our algorithm for obtaining a better accuracy rate by preprocessing the dataset before feed into the model and by proper feature engineering and tuning. The dataset is required to adjust depending on our local market, which is more balanced than this one. Further, a Web tool will be developed for our local market in order to help the local businesses.

\end{document}